\documentclass[sigconf,natbib=true]{acmart}

\copyrightyear{2026}
\acmYear{2026}
\setcopyright{cc}
\setcctype{by}
\acmConference[SIGIR '26]{Proceedings of the 49th International ACM SIGIR Conference on Research and Development in Information Retrieval}{July 20--24, 2026}{Melbourne, VIC, Australia}
\acmBooktitle{Proceedings of the 49th International ACM SIGIR Conference on Research and Development in Information Retrieval (SIGIR '26), July 20--24, 2026, Melbourne, VIC, Australia}
\acmDOI{10.1145/3805712.3809848}
\acmISBN{979-8-4007-2599-9/2026/07}

\usepackage{libertine}
\usepackage{booktabs}
\usepackage{multirow}
\usepackage{pifont}
\usepackage{xspace}
\usepackage{enumitem}
\usepackage{balance}
\usepackage[table]{xcolor} 
\usepackage[most]{tcolorbox}

\definecolor{customblue}{rgb}{0.168, 0.364, 0.557}
\colorlet{framegray}{gray!3!white}
\newcommand{\conv}{\texttt{Conv}\textsubscript{\texttt{Avg}}\xspace}
\newcommand{\cosine}{\texttt{Cos}\textsubscript{\texttt{Avg}}\xspace}

\begin{document}

\title{Context Convergence Improves Answering Inferential Questions}


\author{Jamshid Mozafari}
\authornote{Corresponding Author.}
\orcid{0000-0003-4850-9239}
\affiliation{%
  \institution{University of Innsbruck}
 \city{Innsbruck}
  \country{Austria}
  }
\email{jamshid.mozafari@uibk.ac.at}

\author{Bhawna Piryani}
\orcid{0009-0005-3578-2393}
\affiliation{%
  \institution{University of Innsbruck}
  \city{Innsbruck}
  \country{Austria}
  }
\email{bhawna.piryani@uibk.ac.at}

\author{Adam Jatowt}
\orcid{0000-0001-7235-0665}
\affiliation{%
  \institution{University of Innsbruck}
  \city{Innsbruck}
  \country{Austria}
  }
\email{adam.jatowt@uibk.ac.at}



\begin{abstract}
While Large Language Models (LLMs) are widely used in open-domain Question Answering (QA), their ability to handle inferential questions—where answers must be derived rather than directly retrieved—remains still underexplored. This study investigates how the structure and quality of passages influence LLM performance on such questions. We focus on \textit{convergence}, a measure of how effectively sentences (hints) eliminate incorrect answers, as a criterion for constructing passages. Using subsets of the TriviaHG dataset, we form passages by combining sentences with varying convergence levels and evaluate six LLMs of different sizes and architectures. Our results show that passages built from higher-convergence sentences lead to substantially better answer accuracy than those selected by cosine similarity, indicating that convergence captures meaningful relevance for inferential reasoning. Additionally, ordering sentences by descending convergence slightly improves performance, suggesting that LLMs tend to prioritize earlier, information-rich cues. These findings highlight convergence as a practical signal for guiding passage construction and analyzing inferential reasoning behavior in LLMs.
\end{abstract}





\maketitle

\section{Introduction}\label{s:introduction} 

Question Answering (QA) systems~\cite{10.1007/s10115-022-01783-5} aim to return direct responses to natural language questions. Classic paradigms include FactoidQA~\cite{wang2006survey}, DescriptiveQA~\cite{lee-etal-2005-descriptive}, and BooleanQA~\cite{zhang-etal-2024-boolquestions}, while most methods fall into either \textit{extractive} or \textit{generative} QA~\cite{luo-etal-2022-choose}.  

\begin{figure}[t!]
	\centering
	\includegraphics[width=0.9\columnwidth]{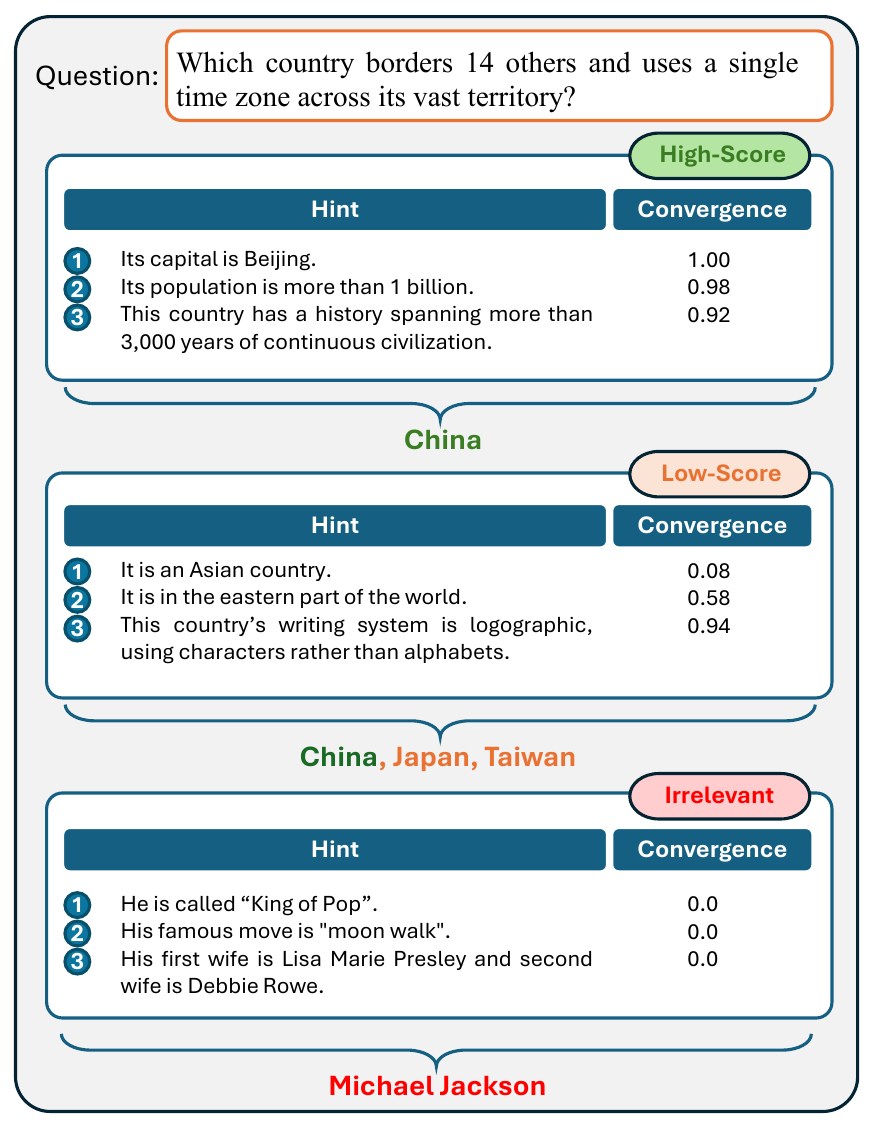}
    \caption{An inferential question accompanied by three distinct sets of hints (treated as passages when concatenated), each associated with convergence scores indicating how strongly the hints align with the correct answer. Each hint set is labeled as \textit{High-Score} or \textit{Low-Score} based on the average convergence of its hints, and as \textit{Irrelevant} if the hints are unrelated to the question. The corresponding answers generated from each set are shown below, where \textcolor{green}{green} denotes the correct answer, \textcolor{orange}{orange} indicates an alternative but plausible answer, and \textcolor{red}{red} represents an incorrect answer.}
	\label{fig:teaser}
\end{figure}

Earlier QA systems were predominantly extractive, relying on encoder-based models such as BERT~\cite{devlin-etal-2019-bert} and RoBERTa~\cite{2019arXiv190711692L} to locate explicit answer spans in passages. With the emergence of decoder-based LLMs like LLaMA~\cite{2024arXiv240721783D} and Gemma~\cite{gemmateam2025gemma3technicalreport}, the field has shifted toward \textit{Retrieval-Augmented Generation (RAG)}~\cite{veturi2024rag} and \textit{Retrieval-Augmented Reasoning (RAR)}~\cite{jin2025search}, driven by their superior generative and reasoning abilities~\cite{minaee2024large}.

Unlike extractive-based models, LLMs are capable of addressing \textit{inferential questions}~\cite{mozafari2026inferentialqa}—questions whose answers are not explicitly stated but must be derived by integrating textual clues, background knowledge, and reasoning. To support such questions, we construct passages from \textit{hints}~\cite{10.1162/tacl_a_00751, 10.1145/3578337.3605119}—concise clues that progressively narrow down the candidate space without revealing the answer directly. A key property of hints is \textit{convergence}~\cite{mozafari2025hinteval}, which evaluates how effectively a hint eliminates incorrect answers and guides reasoning toward the correct one (see Section~\ref{apx:convergence}).

Figure~\ref{fig:teaser} illustrates an inferential question along with three sets of hints—\textit{high-score}, \textit{low-score}, and \textit{irrelevant}—each treated as a passage when concatenated. In the \textit{high-score} set, the average convergence of hints is high, allowing the LLM to generate the correct answer with higher probability. In the \textit{low-score} set, the average convergence is lower, indicating that the hints overlap with other possible answers and may lead the LLM to produce an incorrect output. In the \textit{irrelevant} set, the hints refer to a completely different entity, preventing the LLM from generating the correct answer.


In this work, we study convergence as a principle for selecting and ordering hints (sentences\footnote{From this point onward, we use the term \textit{sentence} instead of \textit{hint}, as passages are composed of multiple sentences. This choice should improve readability for readers.}) to generate passages for inferential questions. We create passages of varying lengths (3–5 sentences) and examine how convergence impacts the ability of LLMs to produce correct answers from them. We further explore how sentence ordering influences performance and conduct experiments across diverse LLMs to assess their capacity for inference beyond extraction. All passages are constructed exclusively from question-relevant sentences; irrelevant hints are not used.

Our findings show that, for inferential questions, passages composed of highly convergent sentences substantially improve answer accuracy over cosine-based selection, confirming convergence as a strong indicator of sentence quality. We further observe that ordering sentences by descending convergence yields slight performance gains, suggesting that LLMs place greater weight on earlier, information-rich cues during inference. To ensure reproducibility, we release the full codebase, datasets, and experimental configurations in a public GitHub repository\footnote{\url{https://github.com/DataScienceUIBK/Context-Convergence-Inferential-QA}}.

In this work, we make the following contributions:
\begin{enumerate}
    \item We introduce and systematically study \textit{convergence} as a novel signal of sentence relevance for passage construction in inferential question answering. Through extensive experiments, we demonstrate that selecting sentences with higher convergence consistently leads to improved answer accuracy compared to conventional cosine similarity-based selection.

    \item We analyze the role of sentence ordering in passage construction and show that arranging sentences in descending order of convergence yields consistent, albeit modest, performance gains. This finding suggests that LLMs exhibit a positional bias, placing greater emphasis on earlier, information-rich cues during inference.
\end{enumerate}

\begin{figure}[t]
\centering
\begin{minipage}{\columnwidth}
\begin{tcolorbox}[
  colback=gray!5,
  colframe=black!70,
  boxrule=0.4pt,
  arc=2pt,
  left=6pt,
  right=6pt,
  top=6pt,
  bottom=6pt,
  fonttitle=\bfseries,
  enhanced,
  title={Prompt for Inferential Questions}
]
\small
\textbf{System:} You are an assistant that answers questions based on the provided context. You just answer questions with exact answers. You do not use sentences as the response.

\vspace{0.5em}
\textbf{User:}
Use the context to answer the question under conditions:

\textbf{1.} Answer should not be sentences. It should be some words. 

\textbf{2.} Do not generate "sorry" or "I cannot ..." sentences; instead, use "NO ANSWER". 

\textbf{3.} Do not generate explanations, reasoning, or full sentences—only provide the exact answer. 

\textbf{4.} If the answer cannot be guessed from the context, respond only with "NO ANSWER".

\vspace{0.5em}
[\textbf{Shot 1}]

\textbf{Context:}
He was the 44th President of the United States. He served as President from 2009 to 2017. He was the first African-American President of the United States. He was a member of the Democratic Party. He was born on August 4, 1961 in Honolulu, Hawaii.

\textbf{Question:} Who won the Nobel Peace Prize in 2009?

\vspace{0.5em}
\textbf{Assistant:} Barack Obama

\vspace{0.8em}

[\textbf{Shot 2}]

\textbf{Context:}
...

\textbf{Question:} 
...

\textbf{Assistant:} ...

\vspace{0.5em}

[\textbf{Shot 3}]
...

[\textbf{Shot 4}]
...

[\textbf{Shot 5}]
...

\vspace{0.8em}
\textbf{Context:}
Its capital is Beijing.  
Its population is more than 1 billion.  
This country has a history spanning more than 3,000 years of continuous civilization.

\textbf{Question:} Which country borders 14 others and uses a single time zone across its vast territory?  

\vspace{0.5em}
\textbf{Assistant:} \textcolor{teal}{China}
\end{tcolorbox}
\end{minipage}
\caption{Few-shot prompt provided to the language model during inference. Each shot demonstrates how to use hint-based context to answer questions concisely with short responses (or \texttt{NO ANSWER} if the context is insufficient). The final pair shows the new query to be answered, with the model-generated answer shown in \textcolor{teal}{green} for clarity.}
\label{fig:full_prompt}
\end{figure}

\begin{table*}[t]
\centering
\caption{Performance (\%) of six LLMs used as QA systems across four evaluation metrics (\textit{ExactMatch}, \textit{Precision}, \textit{Recall}, and \textit{F1}). Each metric is evaluated under the \textit{Convergence} and \textit{Cosine Similarity} conditions for both \textit{High-score} and \textit{Low-score} groups. \textbf{Bold} values indicate the best result for each LLM under a given condition, while \underline{underlined} values denote the overall best score for each LLM across both conditions.}
\label{tbl:exp_all_metrics}

\resizebox{0.85\textwidth}{!}{
\begin{tabular}{l@{\hspace{40pt}}cccccc}
\toprule
\textbf{Metric / Method} & \textbf{LLaMA 3.2 1B} & \textbf{Gemma 3 1B} & \textbf{Qwen 3 4B} & \textbf{Gemma 3 4B} & \textbf{Qwen 3 8B} & \textbf{LLaMA 3.1 8B} \\
\midrule

\multicolumn{7}{c}{\textbf{ExactMatch}} \\ \cmidrule(lr){1-7}
\multicolumn{7}{l}{\textbf{Convergence}} \\
High-score & \textbf{\underline{41.72}} & \textbf{\underline{31.10}} & \textbf{\underline{46.87}} & \textbf{\underline{64.58}} & \textbf{\underline{60.61}} & \textbf{\underline{78.93}} \\
Low-score  & 33.11 & 25.37 & 38.10 & 57.15 & 53.20 & 73.07 \\
\multicolumn{7}{l}{\textbf{Cosine Similarity}} \\
High-score & 36.80 & 26.48 & \textbf{45.45} & \textbf{62.47} & \textbf{60.26} & \textbf{78.25} \\
Low-score  & \textbf{38.29} & \textbf{29.44} & 39.72 & 60.47 & 53.80 & 74.92 \\

\midrule
\multicolumn{7}{c}{\textbf{Precision}} \\ \cmidrule(lr){1-7}
\multicolumn{7}{l}{\textbf{Convergence}} \\
High-score & \textbf{\underline{43.13}} & \textbf{\underline{32.48}} & \textbf{\underline{47.58}} & \textbf{\underline{66.26}} & \textbf{\underline{62.01}} & \textbf{\underline{81.20}} \\
Low-score  & 34.59 & 27.20 & 38.91 & 58.62 & 54.33 & 75.22 \\
\multicolumn{7}{l}{\textbf{Cosine Similarity}} \\
High-score & 38.36 & 28.20 & \textbf{46.55} & \textbf{64.36} & \textbf{61.75} & \textbf{80.47} \\
Low-score  & \textbf{39.37} & \textbf{31.15} & 40.55 & 61.87 & 54.82 & 77.20 \\

\midrule
\multicolumn{7}{c}{\textbf{Recall}} \\ \cmidrule(lr){1-7}
\multicolumn{7}{l}{\textbf{Convergence}} \\
High-score & \textbf{\underline{42.05}} & \textbf{\underline{32.47}} & \textbf{\underline{47.34}} & \textbf{\underline{65.52}} & \textbf{\underline{62.24}} & \textbf{\underline{78.91}} \\
Low-score  & 33.83 & 27.06 & 39.03 & 58.12 & 54.66 & 72.96 \\
\multicolumn{7}{l}{\textbf{Cosine Similarity}} \\
High-score & 37.45 & 28.15 & \textbf{46.24} & \textbf{63.62} & \textbf{62.06} & \textbf{78.27} \\
Low-score  & \textbf{38.59} & \textbf{30.97} & 40.41 & 61.31 & 55.07 & 74.90 \\

\midrule
\multicolumn{7}{c}{\textbf{F1}} \\ \cmidrule(lr){1-7}
\multicolumn{7}{l}{\textbf{Convergence}} \\
High-score & \textbf{\underline{42.11}} & \textbf{\underline{32.02}} & \textbf{\underline{46.84}} & \textbf{\underline{65.24}} & \textbf{\underline{61.55}} & \textbf{\underline{79.33}} \\
Low-score  & 33.85 & 26.77 & 38.45 & 57.82 & 54.02 & 73.41 \\
\multicolumn{7}{l}{\textbf{Cosine Similarity}} \\
High-score & 37.51 & 27.75 & \textbf{45.82} & \textbf{63.37} & \textbf{61.36} & \textbf{78.64} \\
Low-score  & \textbf{38.52} & \textbf{30.66} & 39.97 & 60.99 & 54.47 & 75.37 \\

\bottomrule
\end{tabular}
}
\end{table*}

\section{Related Works}\label{s:related_works}
\paragraph{Question Answering Paradigms}
QA research has evolved from knowledge-based systems~\cite{ ferrucci2010building} to neural approaches. Extractive QA models such as BERT~\cite{devlin-etal-2019-bert} and RoBERTa~\cite{2019arXiv190711692L} identify explicit answer spans, while generative models like T5~\cite{10.5555/3455716.3455856} and LLaMA~\cite{2024arXiv240721783D} synthesize responses. Recent frameworks such as RAG~\cite{lewis-etal-2020-rag, veturi2024rag} and RAR~\cite{jin2025search, shao2025reasonir, liu2025reasonrank, das-etal-2025-rader, long2025diver} combine retrieval and generation to enhance reasoning and factuality.

\paragraph{Inferential and Multi-hop QA}
Multi-hop QA~\cite{yang-etal-2018-hotpotqa, talmor-etal-2019-commonsenseqa} requires aggregating information from multiple supporting facts or documents, often linked through explicit reasoning chains. In contrast, Inferential QA~\cite{mozafari2026inferentialqa} demands reasoning beyond explicit evidence—combining partial clues, commonsense, and background knowledge to derive implicit answers~\cite{7e68810b-3ce0-3862-ae62-c51a27505b15}. While multi-hop tasks emphasize information integration, inferential QA focuses on the deductive process connecting dispersed hints to unseen conclusions.

\paragraph{Hint-based Reasoning and Convergence}
Hint-based QA~\cite{10.1145/3578337.3605119} introduces hints that progressively guide models toward the correct answer. The HintQA~\cite{mozafari-etal-2024-exploring} formalizes this idea by assessing how stepwise hints influence reasoning trajectories and model calibration. However, HintQA primarily evaluates how models adapt across hints, without explicitly modeling the quality or ordering of those hints. In contrast, our work investigates convergence~\cite{mozafari2025hinteval} as a metric for selecting and sequencing hints, examining how the degree of convergence shapes LLM inference accuracy.

\begin{table*}[t]
\centering
\caption{ExactMatch scores (\%) of six LLMs evaluated on passages from \textit{High-score} and \textit{Low-score} groups under two sentence-ordering conditions: \textit{Descending} and \textit{Ascending} (based on convergence scores). \textbf{Bold} values indicate the better ordering for each LLM within the same group, while \underline{underlined} values denote the overall best score for each LLM across both conditions.}
\label{tbl:exp_2_convergence}

\resizebox{0.85\textwidth}{!}{
\begin{tabular}{l@{\hspace{50pt}}cccccc}
\toprule
\textbf{Method} & \textbf{LLaMA 3.2 1B} & \textbf{Gemma 3 1B} & \textbf{Qwen 3 4B} & \textbf{Gemma 3 4B} & \textbf{Qwen 3 8B} & \textbf{LLaMA 3.1 8B} \\
\midrule
\multicolumn{7}{l}{\textbf{High-score}} \\
Descending & \textbf{\underline{36.42}} & \textbf{\underline{26.98}} & \textbf{\underline{39.72}} & \textbf{\underline{56.75}} & \textbf{\underline{54.49}} & \textbf{\underline{71.15}} \\
Ascending  & 35.67 & 25.92 & 39.36 & 56.37 & 53.01 & 71.00 \\
\midrule
\multicolumn{7}{l}{\textbf{Low-score}} \\
Descending & \textbf{28.98} & \textbf{21.81} & \textbf{32.55} & \textbf{50.45} & \textbf{47.57} & \textbf{65.73} \\
Ascending  & 28.55 & 21.49 & 32.02 & 49.94 & 47.53 & 65.24 \\
\bottomrule
\end{tabular}
}
\end{table*}

\section{Convergence}
\label{apx:convergence}

The \textit{convergence} metric captures how effectively a hint reduces the space of plausible answers to a given question. Intuitively, it reflects the degree to which a hint filters out irrelevant or incorrect candidates, thereby steering the reasoning process toward the correct answer. A highly convergent hint provides precise guidance that sharply narrows the candidate set, while a low-convergence hint offers only vague or weak constraints.

To operationalize this notion, we adopt a three-stage procedure inspired by prior work~\cite{10.1145/3626772.3657855,10.1145/3726302.3730299}. First, an LLM generates up to 20 plausible candidate answers for the question, approximating the space of reasonable responses. Second, for each candidate, the model evaluates whether the hint applies to it, producing a binary judgment (\textit{Yes}/\textit{No}). This step determines which candidates remain consistent with the hint.

Finally, the convergence score is computed as the proportion of candidates eliminated by the hint. If the hint is not related to the correct answer, the score is defined as 0, since such a hint is misleading. Otherwise, the score reflects how many incorrect candidates are ruled out relative to the total candidate set. A score of 1 indicates that the hint uniquely isolates the correct answer, whereas a score of 0 indicates that it provides no narrowing effect.

Formally, let $C$ denote the set of generated candidates and let $V \subseteq C$ denote the subset of candidates judged consistent with the hint. The convergence score is defined as:

\begin{equation}
\small
S_{\text{con}} =
\begin{cases}
0, & \text{if the hint is unrelated to the correct answer}, \\
1 - \frac{|V| - 1}{|C|}, & \text{otherwise.}
\end{cases}
\end{equation}

In summary, higher convergence values correspond to hints that strongly align with the correct answer and effectively eliminate competing alternatives, whereas lower values indicate hints that fail to meaningfully constrain the answer space.

\section{Experimental Setups}\label{s:experimental_setups}

To generate passages for inferential questions, we use the TriviaHG dataset~\cite{10.1145/3626772.3657855}, which contains 16,645 questions and 160,230 hints. To the best of our knowledge, only two datasets provide hint-based QA data: TriviaHG and WikiHint~\cite{10.1145/3726302.3730284}. We exclude WikiHint because it includes too few hints per question (on average, only five), making it less suitable for comprehensive experimentation. In contrast, each question in TriviaHG is associated with around nine hints, enabling a more fine-grained analysis of hint selection and ordering.  

For our experiments, we sample 2,000 questions from TriviaHG, selecting those whose hints exhibit a broad range of convergence scores to facilitate a robust analysis of how convergence influences model performance. To access the dataset along with its hints and to compute convergence scores, we utilize the HintEval~\cite{mozafari2025hinteval} toolkit\footnote{\url{https://hinteval.readthedocs.io/}}. For computing convergence scores, we adopt the \textit{LLM-based} variant of the Convergence metric, using LLaMA 3.3 70B~\cite{2024arXiv240721783D} as the underlying model. For cosine similarity computations between sentences, we employ Sentence-BERT~\cite{reimers-gurevych-2019-sentence} with the \textit{all-mpnet-base-v2} model\footnote{\url{https://huggingface.co/sentence-transformers/all-mpnet-base-v2}}.

To evaluate the passages for inferential questions, we employ six LLMs\footnote{We use LLMs with relatively small parameter counts, as larger models often memorize answers internally, which hinders a fair evaluation of passage effects.}: LLaMA~3.2~1B~\cite{2024arXiv240721783D}, Gemma~3~1B~\cite{gemmateam2025gemma3technicalreport}, Qwen~3~4B~\cite{yang2025qwen3}, Gemma~3~4B~\cite{gemmateam2025gemma3technicalreport}, Qwen~3~8B~\cite{yang2025qwen3}, and LLaMA~3.1~8B~\cite{2024arXiv240721783D}. These models span different families and parameter sizes to reduce family-specific biases. We evaluate answer correctness using the \textit{ExactMatch}, \textit{Precision}, \textit{Recall}, and \textit{F1} metrics within the RAG framework to measure how accurately each LLM infers answers from passages.

\section{Experiments and Results}\label{s:experiments_and_results}

\subsection{Convergence Effect on LLMs}\label{ss:effect_of_convergence}

In this experiment, we investigate the effect of \textit{convergence} on sentence selection for constructing passages provided to LLMs. To ensure a fair comparison, it is essential that the passages generated for each question exhibit minimal overlap in the sentences they contain. High overlap could obscure the effect of convergence, making it difficult to determine whether a model’s answer is driven by sentences with higher or lower convergence.

To address this, we create multiple subsets from the nine hints associated with each question, with subset sizes of 3, 4, and 5. Choosing relatively small subset sizes reduces the likelihood of significant overlap between subsets, as each combination covers a different subset of hints, increasing diversity across generated passages. For each subset, we compute the average convergence score (denoted as \conv) and the average cosine similarity between the question and passage's sentences (denoted as \cosine). We then sort all subsets based on their \conv and \cosine values, selecting the ten subsets with the lowest values as the \textit{Low-score} group and the ten subsets with the highest values as the \textit{High-score} group. This procedure is applied separately for convergence and cosine similarity.  

Next, we concatenate the sentences within each subset to form a passage and prompt the LLM to answer the corresponding inferential question based on that passage. The full prompt used for this experiment is shown in Figure~\ref{fig:full_prompt}.

Table~\ref{tbl:exp_all_metrics} presents the results for six LLMs in terms of ExactMatch, Precision, Recall, and F1 metrics, evaluated using both convergence and cosine similarity. 
The results demonstrate that passages with higher \conv scores consistently achieve better performance than those with lower scores. This indicates that convergence is an effective criterion for selecting sentences to construct informative passages for inferential questions. Moreover, the difference in ExactMatch scores between the \textit{Low-score} and \textit{High-score} groups under the convergence metric highlights its strength in distinguishing highly relevant passages from less relevant ones.  

In contrast, the results for \cosine suggest that cosine similarity is not a reliable indicator for sentence selection. This is evident from the inconsistent performance across models—particularly for LLaMA 3.2 1B and Gemma 3 1B—showing that this method does not consistently identify the most relevant passages for inferential questions. Furthermore, the small ExactMatch differences between the \textit{Low-score} and \textit{High-score} groups under \cosine confirm that it struggles to discriminate effectively between high- and low-relevance passages.

Finally, the results show that the \textit{High-score} group under \conv achieves the highest score among all settings, indicating that constructing highly relevant passages for inferential questions requires selecting sentences with the highest convergence. In other words, \textit{passages generated from sentences that provide more detailed and specific information—while covering fewer possible candidate answers—can guide LLMs more effectively toward the correct answer for inferential questions.}

\subsection{Sentence Ordering in Passages}\label{ss:orders_of_hints}

It is important to investigate the effect of sentence order, as several studies~\cite{liu-etal-2024-lost, baker2024lost} have shown that reordering sections or sentences within passages can influence the performance of LLMs and the answers they generate. In our case, the order of sentences within a passage may vary, leading to different versions of the same passage.  

Since the sentences in our passages—originally designed as independent hints—are mutually independent, changing their order does not alter the meaning of the passage. This property enables us to investigate whether LLMs can still answer inferential questions correctly when the sentences are presented in different orders.

In this experiment, we investigate the impact of sentence ordering on LLM performance. We sort the sentences within each set based on their convergence scores in two ways: \textit{Asc} (ascending) and \textit{Desc} (descending). Table~\ref{tbl:exp_2_convergence} shows the results under the \textit{Convergence} condition. We report this condition because, as shown in Section~\ref{ss:effect_of_convergence}, convergence yielded better results than cosine similarity. 

The findings show that ordering sentences in descending order of convergence generally leads to higher EM scores compared to the ascending order. This trend holds for both the \textit{High-score} and \textit{Low-score} groups, although the performance differences are relatively small. This may be due to the short length of sentences—reordering them likely has a limited impact. Nevertheless, the results indicate that order does have an effect.  

Our observations are consistent with prior work~\cite{cuconasu2025rag}, which reported that placing the most relevant passages earlier improves LLM performance. \textit{Our findings extend this conclusion to inferential questions as well, suggesting that LLMs tend to focus more on the earlier and more convergent sentences when generating answers.}

As the final finding from the experiments presented in Sections~\ref{ss:effect_of_convergence} and~\ref{ss:orders_of_hints}, we conclude that the optimal strategy for constructing a passage is to select sentences with the highest average convergence and arrange them in descending order of their convergence scores.

\section{Conclusion}\label{s:conclusion}
This work explored how the structure and content of passages affect LLM performance on inferential questions, focusing on \textit{convergence} as a measure of how strongly sentences guide reasoning toward the correct answer. Experiments across six models show that passages with higher convergence lead to consistently better ExactMatch scores than those selected by cosine similarity, suggesting that convergence effectively captures relevance and reasoning utility.  
We also examined sentence ordering and found that placing highly convergent sentences earlier yields slightly higher performance, implying that LLMs rely more on early, information-rich cues during reasoning.  
Overall, our findings suggest that convergence provides a valuable signal for constructing passages that better support inferential reasoning and for understanding how LLMs process structured contextual information. 

As future work, we plan to replace cosine similarity with stronger retrieval models, such as DPR and ColBERT, to further study the impact of retrieval strategies on passage construction and inferential QA performance.

\begin{acks}
The computational results presented here have been achieved, in part, using the LEO HPC infrastructure of the University of Innsbruck and the Austrian Scientific Computing (ASC) infrastructure.
\end{acks}

\bibliographystyle{ACM-Reference-Format}
\balance
\bibliography{Main}

\end{document}